# Not Color Blind:
# AI Predicts Racial Identity from Black and White Retinal Vessel Segmentations

Aaron S Coyner PhD[1,a], Praveer Singh PhD[2,3,a], James M Brown, PhD[4],
Susan Ostmo MS[1], RV Paul Chan MD[5], Michael F Chiang MD, MA[6],
Jayashree Kalpathy-Cramer PhD[2,3,b], J Peter Campbell MD, MPH[1,b]

***Affiliations***

1. Department of Ophthalmology, Oregon Health & Science University, Portland, OR
2. Radiology, MGH/Harvard Medical School, Charlestown, MA
3. MGH & BWH Center for Clinical Data Science, Boston, MA
4. School of Computer Science, University of Lincoln, Lincoln, UK
5. Ophthalmology and Visual Sciences, University of Illinois at Chicago, Chicago, IL
6. National Eye Institute, National Institutes of Health, Bethesda, MD

[a]Drs. Coyner and Singh contributed to this work equally.
[b]Drs. Kalpathy-Cramer and Campbell supervised this work equally.

***Funding***

This work was supported by grants R01 EY19474, R01 EY031331, R21 EY031883, and P30 EY10572 from the National Institutes of Health (Bethesda, MD), and by unrestricted departmental funding and a Career Development Award (JPC) from Research to Prevent Blindness (New York, NY).

***Conflict of Interest Disclosures***

- Drs. Campbell, Chan, and Kalpathy-Cramer receive research support from Genentech (San Francisco, CA). Dr. Chiang previously received research support from Genentech.
- The i-ROP DL system has been licensed to Boston AI Lab (Boston, MA) by Oregon Health & Science University, Massachusetts General Hospital, Northeastern University, and the University of Illinois, Chicago, which may result in royalties to Drs. Chan, Campbell, Brown, and Kalpathy-Cramer in the future.
- Dr. Campbell is a consultant to Boston AI Lab (Boston, MA).
- Dr. Chan is on the Scientific Advisory Board for Phoenix Technology Group (Pleasanton, CA), a consultant for Alcon (Ft Worth, TX).
- Dr. Chiang was previously a consultant for Novartis (Basel, Switzerland), and was previously an equity owner of InTeleretina, LLC (Honolulu, HI).

## ABSTRACT

### *Background*

Artificial intelligence (AI) may demonstrate racial biases when skin or choroidal pigmentation is present in medical images. Recent studies have shown that convolutional neural networks (CNNs) can predict race from images that were not previously thought to contain race-specific features. We evaluate whether grayscale retinal vessel maps (RVMs) of patients screened for retinopathy of prematurity (ROP) contain race-specific features.

### *Methods*

4095 retinal fundus images (RFIs) were collected from 245 Black and White infants. A U-Net generated RVMs from RFIs, which were subsequently thresholded, binarized, or skeletonized. To determine whether RVM differences between Black and White eyes were physiological, CNNs were trained to predict race from color RFIs, raw RVMs, and thresholded, binarized, or skeletonized RVMs. Area under the precision-recall curve (AUC-PR) was evaluated.

### *Findings*

CNNs predicted race from RFIs near perfectly (image-level AUC-PR: 0.999, subject-level AUC-PR: 1.000). Raw RVMs were almost as informative as color RFIs (image-level AUC-PR: 0.938, subject-level AUC-PR: 0.995). Ultimately, CNNs were able to detect whether RFIs or RVMs were from Black or White babies, regardless of whether images contained color, vessel segmentation brightness differences were nullified, or vessel segmentation widths were normalized.

### *Interpretation*

AI can detect race from grayscale RVMs that were not thought to contain racial information. Two potential explanations for these findings are that: retinal vessels physiologically differ between Black and White babies or the U-Net segments the retinal vasculature differently for various fundus pigmentations. Either way, the implications remain the same: AI algorithms have potential to demonstrate racial bias in practice, even when preliminary attempts to remove such information from the underlying images appear to be successful.

## INTRODUCTION

The effect of conscious and unconscious bias in medicine has rightly become an important area of research, education, and advocacy. It is well understood that human clinicians may be biased by patients' gender, sexual orientation, weight, race, and/or ethnicity and that this may affect diagnoses and subsequent treatment decisions.[1–6] With the number of applications of artificial intelligence (AI) algorithms in clinical medicine increasing, so has the attention paid to the potential of these algorithms to be biased. Since AI algorithms may be deployed at higher scale and across more populations than a single clinician, the potential for transferred harm from biased algorithms is greater.[7]

Regarding race, it has been well documented that images captured using the visible-light spectrum, such as skin or retinal fundus images (RFIs), contain information related to pigmentation.[6,8–10] Aside from previous research demonstrating that mammogram breast density is predictive of race, it has been assumed that non-visible-light imaging modalities, such as X-rays, computed tomography (CT) scans, magnetic resonance images (MRIs), optical coherence tomography (OCT) scans, etc., do not capture such information.[11,12] However, Banerjee *et al.* recently demonstrated that deep learning (DL), a subcategory of AI, could detect patients' self-reported race from X-rays and CT scans.[11] Specifically, they trained multiple deep convolutional neural networks (CNNs) to detect race from three different datasets of chest X-rays, three different datasets of chest CT scans, one dataset of hand X-rays, one dataset of breast mammograms, and one dataset of lateral C-spine X-rays. All DL models were able to detect patients' self-reported race with a high degree of certainty. Furthermore, images were distorted and manipulated in ways that made them, at times, entirely undiagnosable by human experts, but AI was still able to accurately predict race using them.

In ophthalmology, there exist many different types of imaging modalities, such as retinal photography, OCT, slit lamp photography, scanning laser ophthalmoscopy (SLO), etc., and many of these modalities use visible light.[6,7,13] As useful as these modalities and their associated images are for diagnosis, they may be problematic for AI, as it has been shown that there is a high degree of correlation between race and the color of the fundus oculi (**Figure 1**).[6,7,10,14] In part to circumvent this issue, AI researchers have employed algorithms for automated segmentation of the retinal vasculature to create grayscale retinal vessel maps (RVMs), under the assumption that "this process effectively eliminates variations in pigmentation [and] illumination."[15–17]

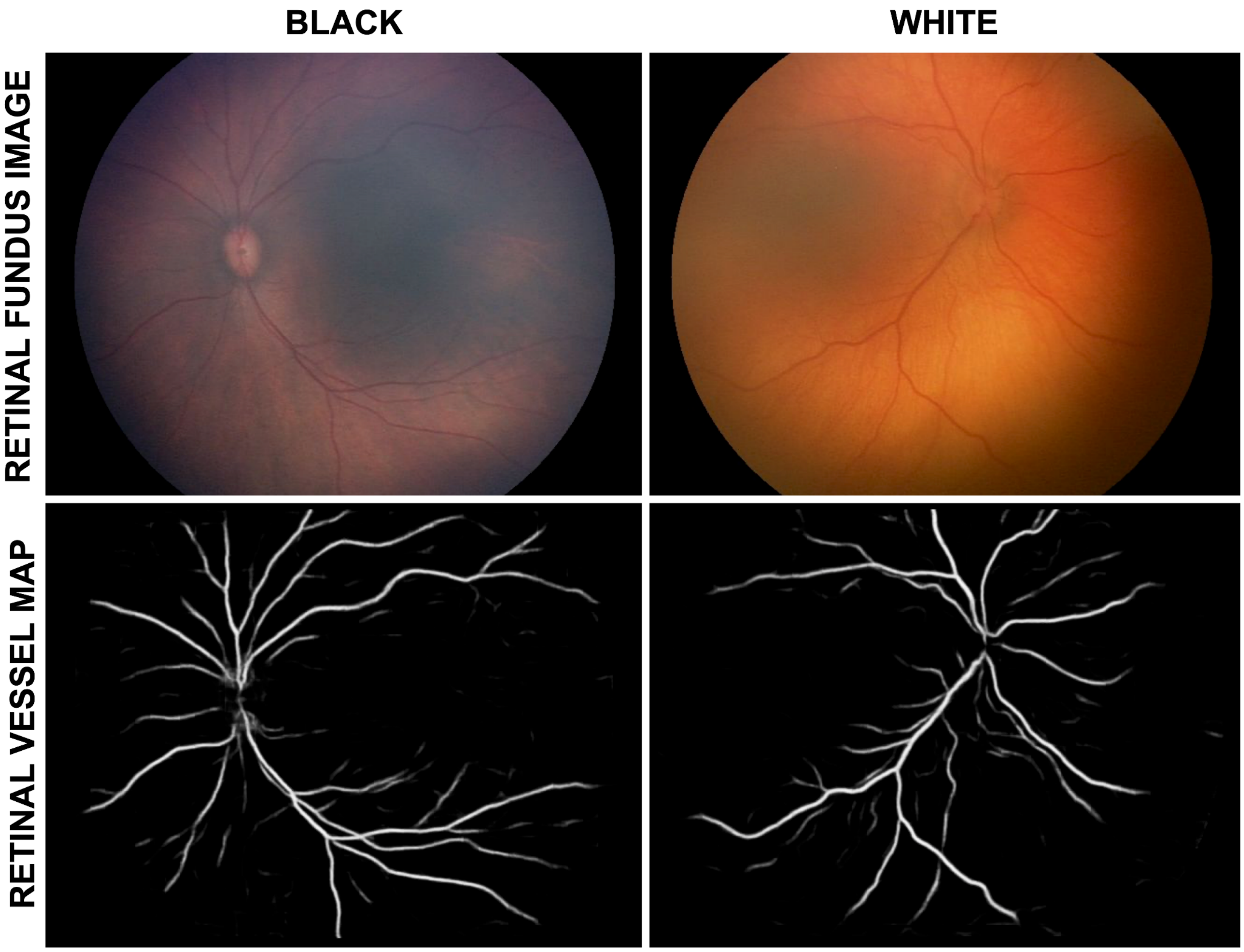

**Figure 1: Example retinal fundus images (RFIs) and associated retinal vessel maps (RVMs) collected from Black and White babies.** *Top*: RFIs were collected via (visible-light) retinal photography from premature infants. In general, Black babies (*left*) tend to have more darkly pigmented eyes than White babies (*right*). *Bottom*: Corresponding RVMs that were segmented from RFIs (*top*) using a U-Net. This method, theoretically, removes pigmentation and luminance information contained within RFIs.

Herein, we test this hypothesis using a dataset of RFIs obtained via retinal photography from infants who were screened for a potentially-blinding disease, retinopathy of prematurity (ROP). Treatment-requiring ROP (TR-ROP) has traditionally been identified by examining infants at the bedside via ophthalmoscopy.[18] However, recent advances in imaging, telemedicine, and AI have allowed for both remote and automated screening and diagnosis.[15,18–20] Automated methods for the diagnosis of ROP (and other retinal vascular diseases) have used RVMs as input, rather than RFIs, since it is believed that they do not contain information related to pigment or nonvascular pathologies; thus, they theoretically only allow the algorithm to learn information related to retinal vessel dilation and tortuosity.[15,16,21] In this study, we evaluate whether AI can detect parent-reported race from RFIs, associated RVMs, as well as thresholded, binarized, and skeletonized versions thereof.

## METHODS

### *i-ROP Study*

The Imaging and Informatics in ROP study (i-ROP) was approved by the Institutional Review Boards at the coordinating center (Oregon Health & Science University) and at each of seven study centers (Columbia University, University of Illinois at Chicago, William Beaumont Hospital, Children's Hospital Los Angeles, Cedars-Sinai Medical Center, University of Miami, Weill Cornell Medical Center) and was conducted in accordance with the Declaration of Helsinki.[22] Written, informed consent was obtained from parents of all enrolled infants.

### *Dataset*

As part of the multicenter i-ROP cohort study, conducted between January 2012 and July 2020, subjects born prior to a gestational age (GA) of 31 weeks or with a birth weight (BW) less than 1501 grams were routinely screened for ROP. During each exam, five RFIs (nasal, temporal, inferior, superior, and posterior pole-centered views) were captured for each eye via a RetCam (Natus; Pleasanton, CA). Subjects were clinically examined at the bedside, but also received image-based ROP diagnoses, which were determined by a consensus of three ROP experts using the full International Classification of ROP (ICROP) criteria.[18] To be included in this study, subjects were required to meet the inclusion criteria presented in **Table 1**.

**Table 1: Subject inclusion criteria.**

| Inclusion Criteria | Determined By |
|---|---|
| Birth Weight < 1501 grams | Examining Physician |
| Gestational Age < 31 weeks | Examining Physician |
| Non-Hispanic | Parent-reported |
| Single-race | Parent-reported |
| Black or White race | Parent-reported |
| Normal retinal vasculature | ROP Expert Consensus |
| Never treated for ROP | ROP Expert Consensus |
| No siblings in the dataset | Study Coordinator |

### *Partitioning*

Using the *groupdata2* package in R, the 245 infants and their associated RFIs and RVMs were partitioned into training, validation, and test datasets using a 50/20/30 split by subject identification number.[23,24] Thus, subjects were exclusive to the training, validation, or test datasets, and the natural distribution of race was preserved as best as possible. Furthermore, statistical differences between BW, GA, and postmenstrual age (PMA), across datasets, were assessed via Welch's two-sample t-test (significance defined as $p \leq 0.05$).

***Retinal Vessel Map Modification***

All color RFIs were segmented into grayscale RVMs using a previously-trained U-Net, which predicts, for each pixel in an image, a probability of that pixel representing a main artery or vein (i.e., non-zero pixel values indicate that there is at least some probability of said pixels representing a main artery or vein).[15,21] The dataset was also transformed into multiple training, validation, and test datasets (using the same splits as above) based on transformations of the original grayscale RVMs via thresholding, binarizing, or skeletonizing.

*Thresholding*

RVMs were thresholded — in this context, pixel values below a given threshold were set equal to 0, those above the threshold were unchanged — at pixel intensity values (PIVs) of 0 (i.e., "raw" U-Net segmentations/RVMs), 50, 100, 150, 200, 210, 220, 230, 240, 250, and 256. These PIV thresholds corresponded to the U-Net's segmentation confidence/vessel probability being equal to or greater than 0.0%, 19.6%, 39.2%, 58.8%, 78.4%, 82.4%, 86.32%, 90.2%, 94.1%, 98.0%, and 100%, respectively. These values were calculated by normalizing the PIVs (vessel probability = PIV/255). This provided 11 levels of confidence thresholds to attempt to subtract any race information that was leaking through into vessel segmentations. In doing so, we were able to constrain the available pixel intensity values in RVMs.

*Binarization*

Binarization was performed on thresholded image sets. For each threshold, RVMs were binarized by converting all pixels with a non-zero probability of belonging to a vessel to 100% (i.e., PIV was set equal to 255). This not only helped to visualize the pixel-level information present in the U-Net RVMs, but also to remove pixel intensity differences of segmented arteries and veins that may occur between Black and White babies.

*Skeletonization*

Skeletonization was performed on binarized versions of thresholded RVMs. Binarized images were further transformed into retinal vessel skeletons using a method described by Lee *et al*.[25] This algorithm reduced the width of each vessel to a single pixel and effectively negated any differences that may have existed between segmented vessel widths in Black versus White RVMs.

*Low and Mid-Range*

Additionally, two other sets of images were created: (1) a set of images where pixel values *above* 10 were zeroed, resulting in seemingly-black images to the human eye, and (2) a set of images where pixel values below 75 or above 150 were zeroed. These sets of images were created to test whether race information was also contained only in the low- and high-range of pixel intensity values or in the mid-range as well. These sets of images were also binarized and skeletonized.

***Model Training***

Multiple ResNet-18 CNNs were trained for binary classification using the Python PyTorch library.[26–28] Specifically, CNNs were trained to detect Black versus White parent-reported race using color RFIs, or each iteration of thresholded, binarized, and skeletonized RVMs. Batches of 64 images, with random flips and rotations applied, were fed into CNNs at a size of 224x224 pixels. Using stochastic gradient descent, models were allowed to train for up to 10 epochs at a constant learning rate of 0.001. However, early stopping was implemented, with cessation of model training occurring if validation loss did not improve after five epochs.

***Model Evaluation***

Due to the imbalance between Black and White babies in the dataset, area under the precision recall curve (AUC-PR) was used as the main metric of model performance, however area under the receiver operating characteristics curve (AUC-ROC) was also evaluated. After models finished training, the probabilities of images in the test dataset belonging to Black or White babies were predicted. Since subjects could have multiple images (various views and exams), subject-level predictions were determined using the median probability value for all images belonging to a subject. AUC-PR and AUC-ROC were evaluated at both the image-level and subject-level.

***Inherent Retinal Fundus Image and Retinal Vessel Map Differences***

To explore whether inherent differences between RFIs of Black versus White eyes existed, histograms of the red, green, and blue channels were developed from the training dataset. For RVMs, the number of pixels segmented, between Black and White babies' RVMs, was investigated. This was performed by counting the number of non-zero pixels for thresholded/binarized RVMs and skeletonized RVMs in the training dataset at a threshold of 0 (i.e., raw RVM outputs from U-Net), and at 50 and 200 to remove superfluous vessel information/"noise".

***Repeatability***

The experiments mentioned above were performed by ASC. However, to verify that data leakage (i.e., subject crossover between datasets) or method-related issues (e.g., methods for training models, evaluating performance, etc.) were not of concern, a subset of these experiments were repeated by PS (**Supplemental Methods**).

## RESULTS

### *Partitioning*

Of the babies in the i-ROP dataset, 245 (94 Black/151 White) met inclusion criteria for this study (**Table 1**). These babies had a total of 4095 RFIs (1565 Black/2530 White) collected from them that were available for training, validating, and testing CNNs. After partitioning the data into training, validation, and test datasets — ensuring that subjects were mutually exclusive to each dataset and that the natural distribution of race remained — significant differences between BW, GA, and PMA were assessed. The only significant difference ($p \leq 0.05$) between the two groups was the PMA of the babies in the training dataset (**Table 2**). Notably, the demographics of the two groups were similar and typical of ROP screening populations, and none of the included subjects had abnormal retinal vasculature, as determined by a consensus of ROP experts.

**Table 2: Dataset characteristics.**

| Metric | Black | White | *p*-value |
|---|---|---|---|
| ***Training Dataset*** | | | |
| Birth Weight ± SD | 1008.0 ± 340.0 | 1061.1 ± 320.3 | 0.854 |
| Gestational Age ± SD | 27.7 ± 2.3 | 27.6 ± 2.3 | 0.405 |
| Postmenstrual Age ± SD | 35.5 ± 4.7 | 34.4 ± 2.9 | **0.028** |
| Number of Images (%) | 754 (38.2) | 1221 (61.8) | — |
| Number of Subjects (%) | 45 (39.1) | 70 (60.9) | — |
| ***Validation Dataset*** | | | |
| Birthweight ± SD | 882.6 ± 277.2 | 1030.7 ± 289.5 | 0.151 |
| Gestational Age ± SD | 26.8 ± 2.6 | 27.8 ± 2.0 | 0.090 |
| Postmenstrual Age ± SD | 35.0 ± 3.1 | 34.5 ± 2.4 | 0.315 |
| Number of Images (%) | 288 (37.4) | 482 (62.6) | — |
| Number of Subjects (%) | 18 (39.1) | 28 (60.9) | — |
| ***Test Dataset*** | | | |
| Birthweight ± SD | 1096.9 ± 318.0 | 1098.5 ± 305.2 | 0.875 |
| Gestational Age ± SD | 27.9 ± 2.3 | 28.0 ± 2.3 | 0.983 |
| Postmenstrual Age ± SD | 34.7 ± 2.5 | 34.4 ± 2.4 | 0.466 |
| Number of Images (%) | 523 (38.7) | 827 (61.3) | — |
| Number of Subjects (%) | 27 (38.6) | 43 (61.4) | — |
| ***All Data*** | | | |
| Birthweight ± SD | 1013.7 ± 324.9 | 1059.5 ± 308.2 | 0.731 |
| Gestational Age ± SD | 27.6 ± 2.3 | 27.7 ± 2.2 | 0.276 |
| Postmenstrual Age ± SD | 35.1 ± 3.8 | 34.6 ± 2.8 | 0.067 |
| Number of Images (%) | 1565 (38.2) | 2530 (61.8) | — |
| Number of Subjects (%) | 94 (38.4) | 151 (61.6) | — |

***Retinal Vessel Map Modification***

As expected, increasing the threshold at which pixels below a threshold were zeroed resulted in increasingly darker images (i.e., less of the retinal vasculature was segmented since the probability threshold for U-Net determining whether a pixel represented a vessel was increased; **Figure 2**). Binarized RVMs allowed for visualization of all pixel values with a non-zero probability of belonging to retinal blood vessels, as determined by the U-Net. When the threshold was set to 0 (i.e., the raw RVM outputs from U-Net), the binarized output revealed that many pixels not belonging to the main retinal vascular tree were segmented (**Figure 2, middle row, panel "PIV $\geq$ 0"**). At the other extreme, as expected, the appearance of the retinal vasculature in grayscale RVMs began to fade as the threshold at which pixels below a certain value were zeroed out increased. Vessels became harder to visualize at a threshold equal to or greater than 200 (**Figure 2, panel "PIV $\geq$ 200"**). At a threshold of 250, very few pixels (if any) were present. At a threshold of 256, all pixel values were 0; thus, RVMs were completely black.

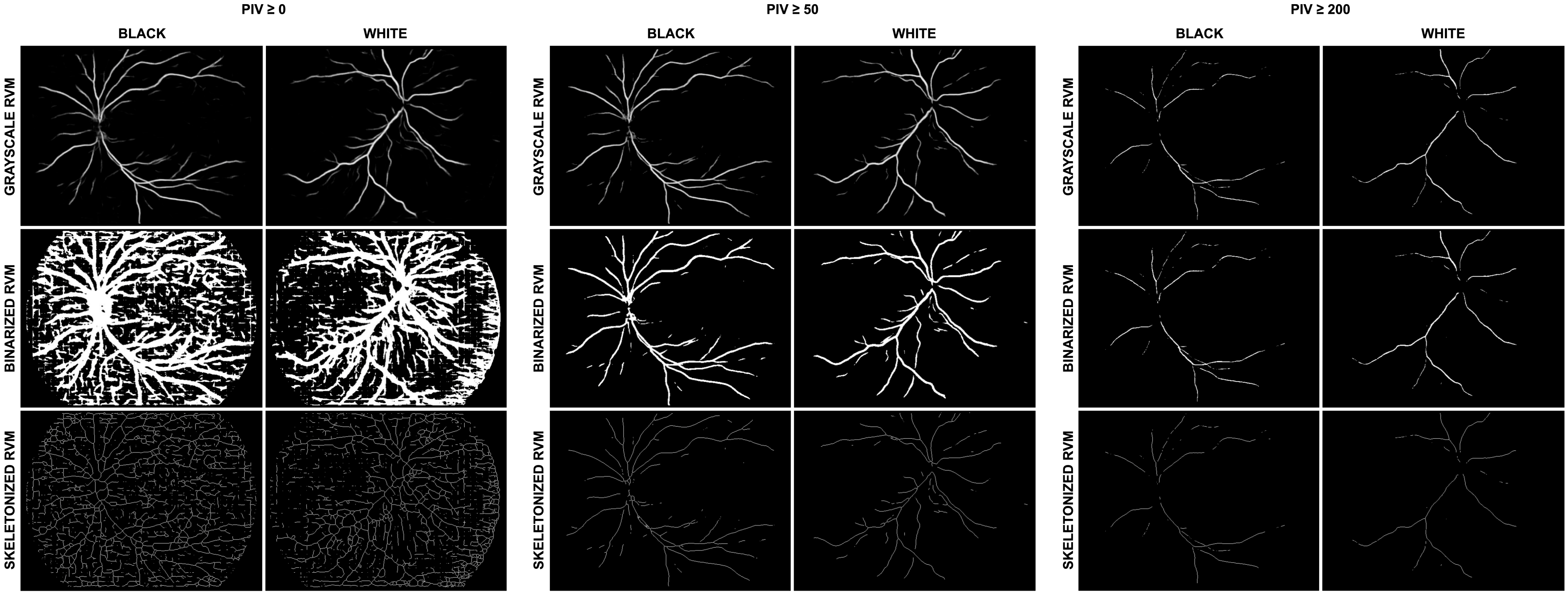

**Figure 2: Example retinal vessel maps (RVMs) and modified RVMs.** Grayscale RVMs (top row) were generated, using a U-Net, from the RFIs in Figure 1. Pixel values below the set thresholds (left, middle, and right panels) were zeroed out (top row), binarized (middle row), and skeletonized (bottom row). Models were trained and assessed on all types of images across all thresholds.

***Prediction of Race Using Retinal Fundus Images and Retinal Vessel Maps***
A total of 40 different ResNet-18 models were trained on color RFIs, or thresholded grayscale, binarized, or skeletonized RVMs. Test dataset AUC-PR and AUC-ROC, at both the image level and subject level for each model, are presented in **Supplemental Table 1**. We found that ResNet-18 models had near perfect ability to predict parent-reported race from color RFIs, with an AUC-PR of 0.999 at the image level and 1.000 at the subject level. Interestingly, raw (threshold equal to 0) grayscale RVMs were nearly as predictive as color RFIs (image-level AUC-PR: 0.938, subject-level AUC-PR: 0.995). That is, the CNN was able to predict Black versus White race using grayscale vessel maps that, to the human eye, contain no appreciable information regarding fundus pigmentation. Similar results were found by another researcher in our group (PS, **Supplemental Methods, Supplemental Table 2**).

To further investigate this, we iteratively trained models using less and less information through a process of thresholding, binarizing, and skeletonizing. As the threshold at which pixels below the threshold were zeroed increased, the average image- and subject-level AUC-PR for grayscale, binarized, and skeletonized images decreased. However, even at the upper end of thresholding, models retained better-than-chance predictive ability (**Figure 3, Supplemental Table 1**).

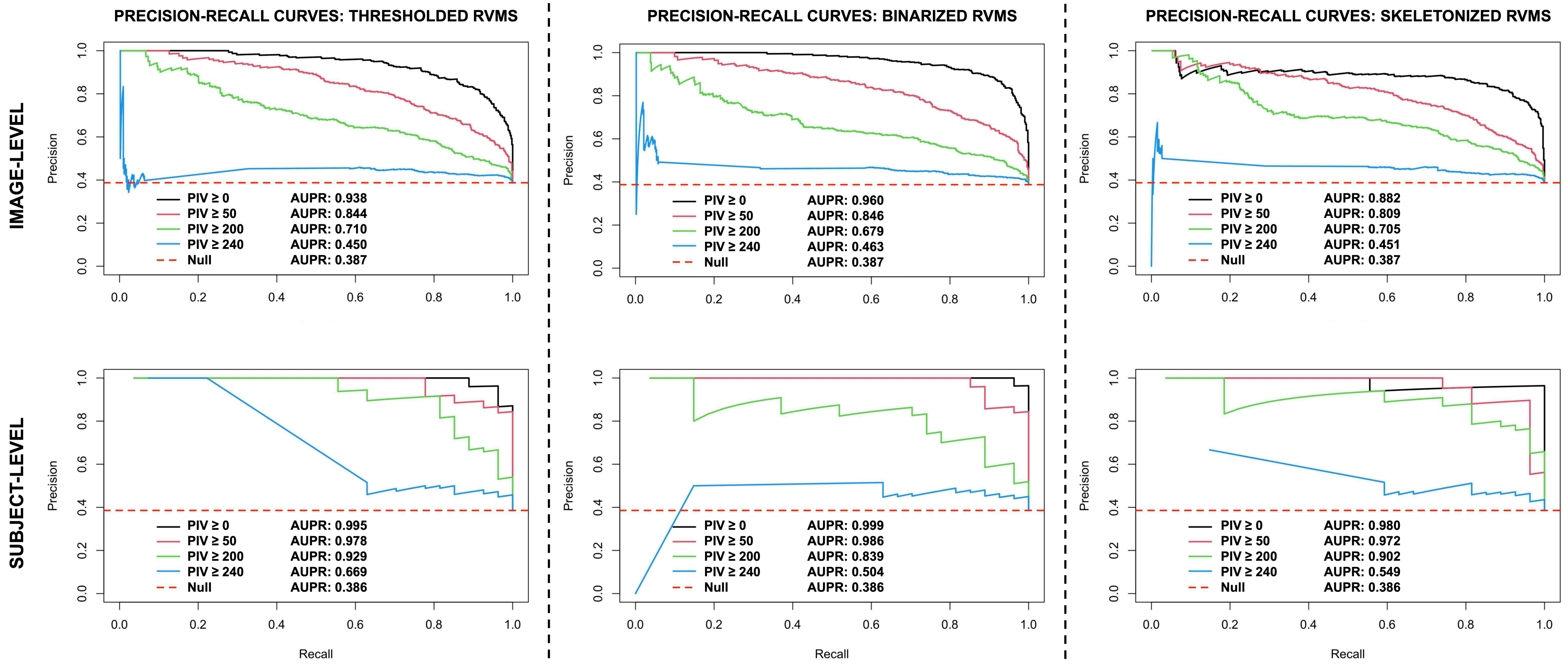

**Figure 3: Image- and subject-level precision-recall curves for retinal vessel maps (RVMs) across select thresholds.** Image-level (top) and subject-level (bottom) precision-recall curves for thresholded (left), binarized (middle), and skeletonized (right) RVMs. In general, as the threshold at which pixel values are zeroed increases, performance of the models decreases.

At the low-threshold end of the spectrum, models trained on RVMs with pixel values greater than 10 zeroed retained predictive utility with subject-level AUC-PRs ranging from 0.945 to 0.994, despite having no pixels visible to the human eye in the grayscale RVMs (**Figure 4, Supplemental Table 1**). In the mid-threshold case, with PIVs less than 75 and greater than 150 zeroed, models retained subject level AUC-PRs ranging from 0.972 to 0.973, despite being constrained to the mid-range of PIVs (**Figure 4, Supplemental Table 1**).

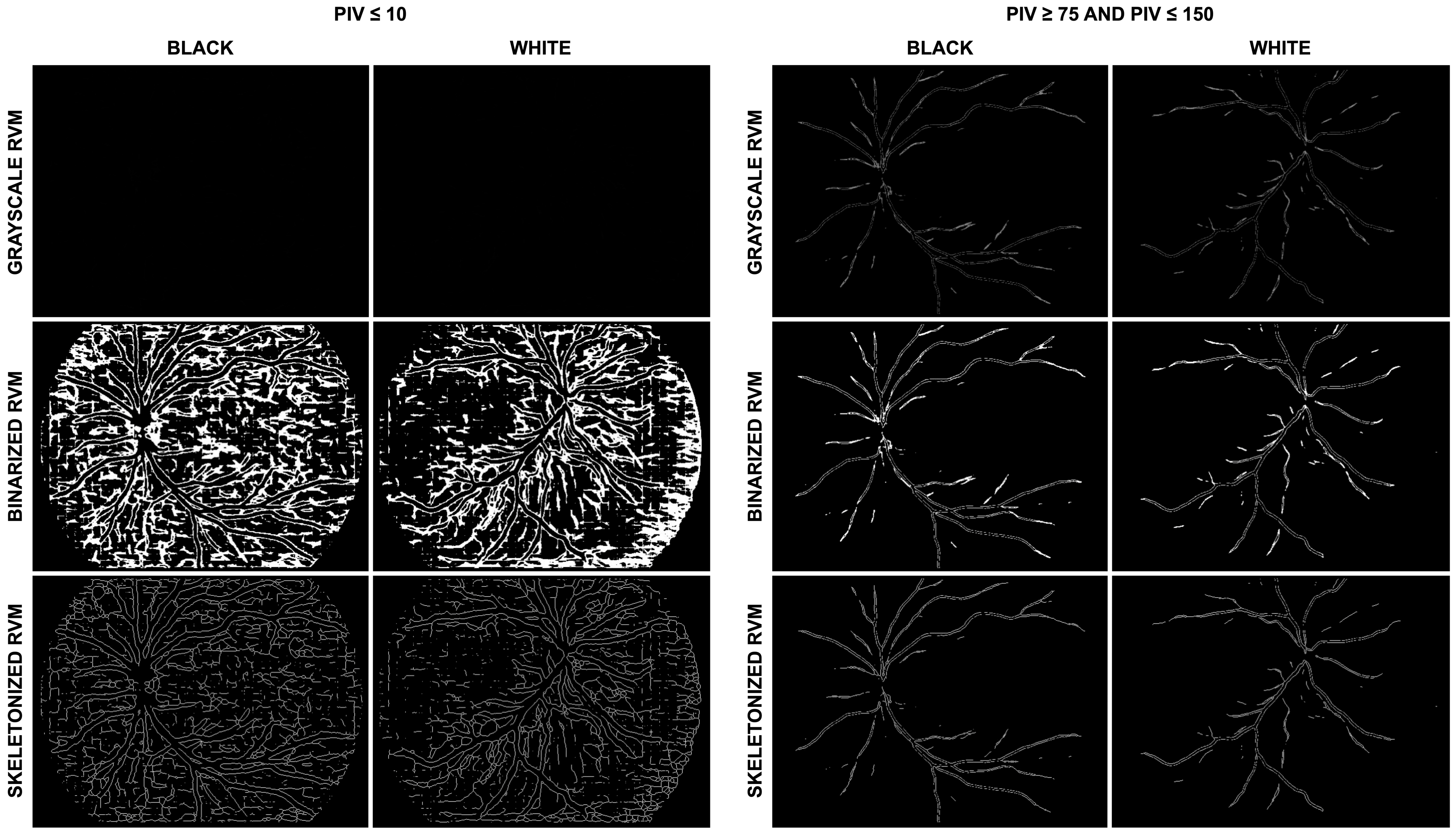

**Figure 4: Example retinal vessel maps (RVMs) thresholding pixel intensity values (PIVs) greater than 10 or outside of inner bounds.** In addition to previous thresholding, PIVs greater than 10 (left) or PIVs less than 75 and greater than 150 (right) were zeroed. Binarized RVMs (middle row) revealed that much information was still contained within RVMs containing only PIVs less than 10. Vessel skeletons are presented in the bottom row.

### ***Inherent Retinal Fundus Image and Retinal Vessel Map Differences***

To investigate whether inherent RFI differences between Black and White babies' eyes existed, histograms of the red, green, and blue channels were evaluated. In general, White babies had higher intensities of red present in their RFIs, as opposed to Black babies (**Figure 5, "Red Channel"**). Similarly, Black babies had higher intensities of blue present in their RFIs (**Figure 5, "Blue Channel"**). In theory, a CNN could easily learn these features and use them to predict race.

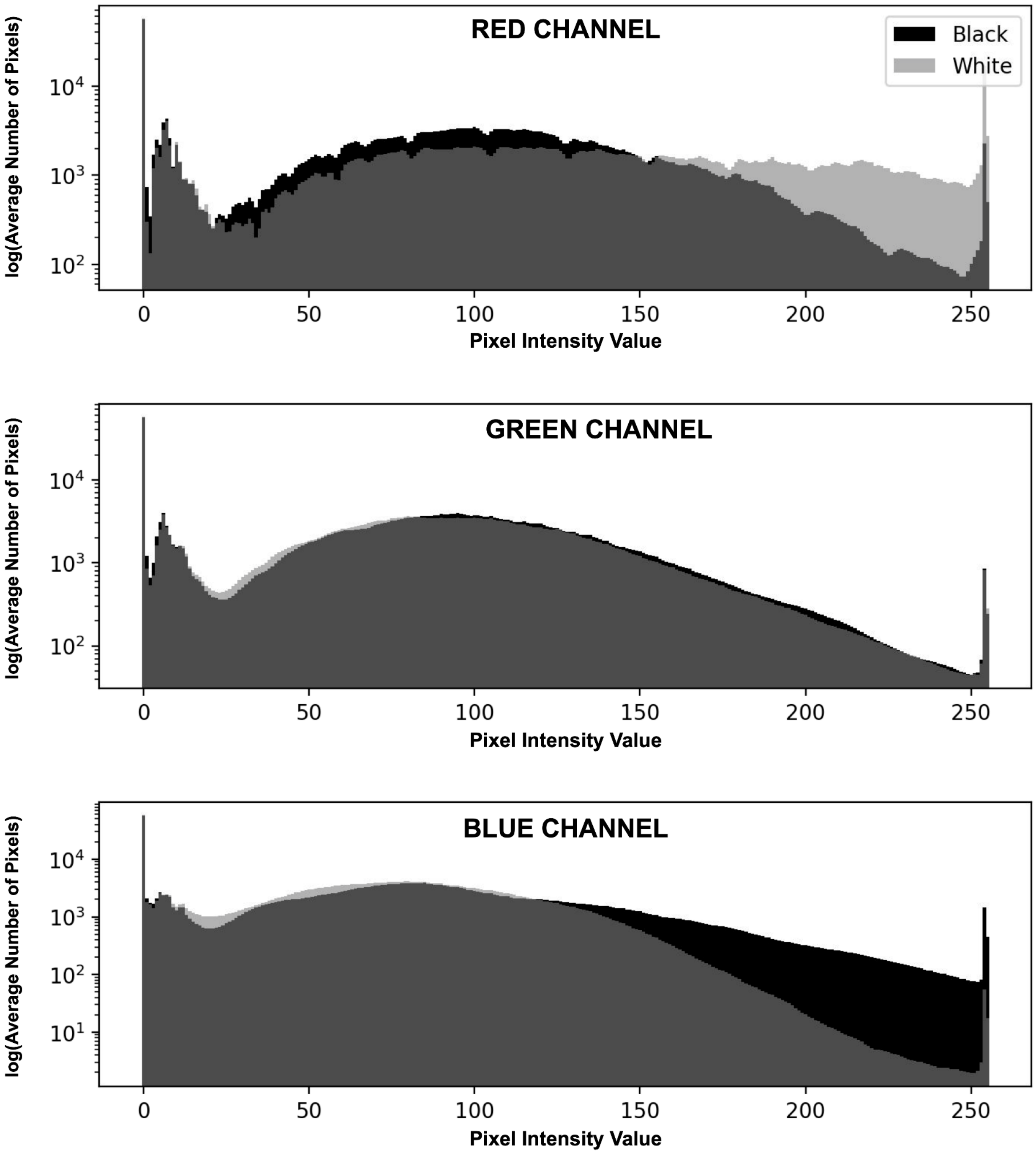

**Figure 5: Histograms of color retinal fundus images (RFIs).** Histograms of the red (*top*), green (*middle*), and blue (*bottom*) channels of color RFIs were generated. White babies had higher intensities of red present in their retinal fundus images, whereas Black babies had higher intensities of blue. Green intensities seemed to be equal among races.

RVMs do not contain any color, therefore a CNN could not use that information to predict race. However, the number of pixels segmented (i.e., the number of pixels with *some* probability of belonging to a main artery or vein) was evaluated. Grayscale and skeletonized RVMs without any thresholding applied displayed distinct differences in this regard (**Figure 6, Top Panel**). Thresholding these images, however, increased the overlap between number of pixels segmented for Black and White babies' RVMs (**Figure 6, Middle and Bottom Panels**). Interestingly, thresholded skeletonized RVMs did not completely overlap, suggesting that more major arteries and veins are segmented for White eyes than Black eyes by the U-Net.

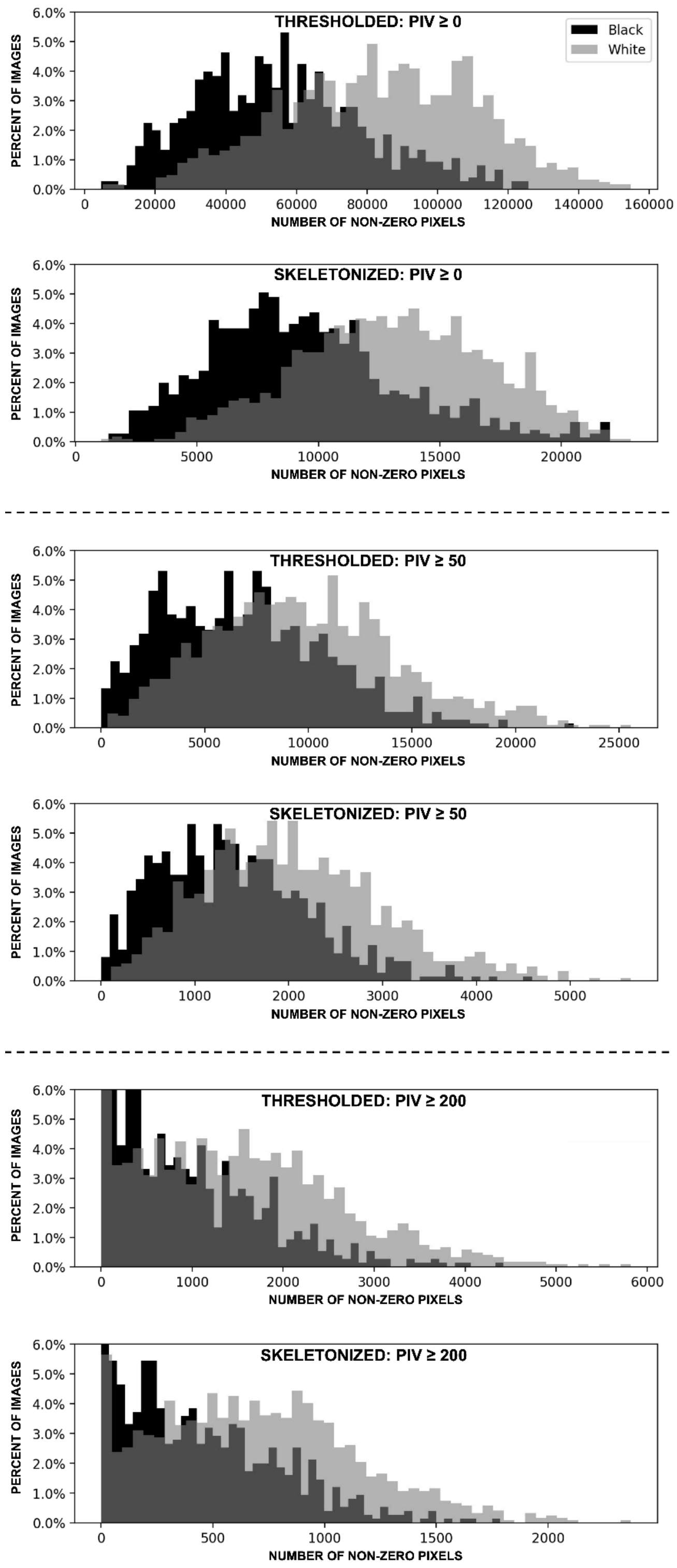

**Figure 6: Number of pixels segmented by the U-Net before and after thresholding and skeletonizing.** The number of pixels segmented in raw RVMs and skeletonized version thereof are distinctly different between Black and White eyes (*top*). Increased thresholding of these RVMs results in far better overlap, but differences still exist (*middle, bottom*). Removing segmented vessel width via skeletonizing still resulted in differences between groups.

## DISCUSSION

We found that AI was easily able to predict the race of babies from retinal vessel segmentations that contain no visible information regarding pigmentation. As we modified the grayscale images to narrow the pixel intensities/vessel probabilities via thresholding, model performance began to decrease, as expected; however, even images that appeared devoid of information to the naked eye retained predictive information of the race of the original baby. There are two potential explanations for these non-intuitive findings, and one important implication. Either there are true differences in the appearance of retinal vessels between Black and White babies or the way U-Net segments vessels is biased due to the underlying differences in a perceived appearance. Either way, this finding has implications for deployed algorithms that use grayscale segmentations to, theoretically, mitigate the risk of racial bias.

One possibility is that the retinal vessels of Black babies appear different from the vessels of White babies in some way that AI can appreciate, but humans cannot. Several previous studies have looked at this general topic using various methodologies.[29–34] Rochtchina *et al*. reported that the retinal vascular caliber of children of East Asian descent was significantly wider than in Caucasian children; they concluded that this effect was likely due to contrast and ability of human graders and AI to detect the edges of the vessels.[30] However, Li *et al*. performed a study on a multiethnic Asian population (Chinese, Malay, and Indian; i.e., all eyes were similarly pigmented) free of clinical diseases and found that there were significant differences between retinal vascular caliber, fractal dimension, and tortuosity between ethnicities.[29] Similarly, Wong *et al.* showed, after adjusting for age, gender, and imaging site, that Black eyes were found to have significantly larger arteriolar and venular calibers than White eyes.[31] However, even if this is true, it is unlikely that the explanation for such high model performance on RVMs is simply due to vessel caliber, since skeletonized (single-pixel wide) RVMs retained predictive ability. That said, we cannot exclude the possibility that higher order feature relationships exist in these data.

Our leading theory is that the U-Net itself is biased in how it determines the probability of pixels belonging to parts of retinal vessels. One possible explanation is that the U-Net was trained using a dataset of predominantly White babies (93%).[15] This could result in differences in the U-Net's ability to distinguish retinal from choroidal vessels, the ability to determine the vessel borders, or the ability to segment the entire lengths of the vessel segments to the borders of the image in darker fundi. This is supported by the finding that the histograms of pixels segmented have higher tails in White babies, suggesting that more vessels are segmented overall in lighter eyes (**Figure 6**). It is not clear whether this is due to the raw ability of the camera resolution to identify vessel segments or the ability of the U-Net to distinguish vessel versus non-vessel, but the end result is the same. As we zeroed out pixel values below, above, and between specific thresholds, both the image- and subject-level AUC-PRs began to drop; even highly thresholded and skeletonized vessel maps retained some predictive information regarding race. However, the U-Net was trained on RFIs that were first converted to grayscale images and subjected to contrast adjustment — specifically, contrast limited histogram equalisation (CLAHE) — and was therefore never actually trained on color RFIs.[15] Thus, we are as of yet unable to fully explain these findings based on the U-Net hypothesis alone.

Furthermore, these results do not appear to be related to differences in demographics, disease prevalence, or treatment differences by race. We ensured that included babies were currently free of ROP (i.e., their retinal vasculature appeared to be growing normally and was not dilated or tortuous), as determined by three ROP experts, and excluded any babies that were previously treated for ROP.[18,35] This was an important consideration, since there may exist (unintended) differences in treatment protocols for different races, as has been witnessed in other fields.[1,2,4,5] Finally, differences between BW, GA, and PMA were evaluated, as it was possible that there could exist differences in vascular appearance between older and/or heavier babies as compared to their counterparts. The only significant difference occurred for the PMA between Black and White babies in the training dataset, however this was not the case in the validation or test datasets and should therefore not have an effect on the results reported in this study. We also excluded any babies who had siblings in the original dataset, to avoid any potential similarities due to genetics. Ultimately, neither an underlying disease distribution nor clinical factors should be confounders in this dataset or study.

Similar to the recent work by Banerjee *et al.*, the key message here is not that AI models can detect race from grayscale, binarized, or skeletonized retinal images, but that because they can, there exists a risk of racial bias in medical AI algorithms that use them as input.[11] It is safe to assume that these results and concerns should extend outside of ROP and into other ophthalmic diseases that use RFIs or RVMs. Additionally, because Banerjee *et al.* were able to demonstrate that radiology-based non-visible-light imaging modalities retained information regarding race, investigation into ophthalmology-based non-visible-light imaging modalities, such as OCT and SLO, should also be investigated for this bias. Again, even though there may not be biomarkers of race visible to humans in these types of images, that does not mean they do not exist. In addition to previous demonstrations that X-rays and CT scans contain information regarding race, we have now shown that RFIs and, unexpectedly, RVMs also contain this information.[11]

It should be noted that potentially "intermediate" groups were excluded from this study (e.g., Hispanic ethnicity, Asian race, etc.). These groups may be thought of as intermediate in pigmentation, which could prove more difficult for an algorithm to identify than the Black versus White eyes. Future work will investigate the ability of DL to detect these eyes, both to test the hypothesis that it is possible and also to reduce the spectrum bias present in this study (i.e., participants in this study were at the “extremes” of the pigmentary spectrum).

In this paper, AI models trained on thresholded, binarized, and skeletonized grayscale retinal vessel maps could identify parent-reported race with extremely high accuracy, which is not something that human graders can do. This could not be fully explained by modifiable differences in vessel segmentations in this study, although that remains a possibility. Luckily, we have demonstrated that ROP detection models that rely on these particular U-Net segmentations for disease detection perform well in various populations.[19,36] However, as increased attention is being paid to the potential for racial bias in medical AI algorithms, one proposed strategy in retinal fundus images has been to focus on interpretable pre-processed features, rather than an entirely black box strategy, to reduce the potential for exactly this sort of bias. These results suggest that these preliminary steps may not always be successful, and future work ought to ensure as much attention is paid to potential sources of bias in the pre-processing steps as in the training of the final diagnostic models.

## CODE AND DATA AVAILABILITY

Code for this study is available at https://github.com/aaroncoyner/not-color-blind. Data will not be available due to patient privacy considerations.

## SUPPLEMENTAL TABLES

**Supplemental Table 1: Performance of models.**

| IMAGE TYPE | AUC-PR (image level) | AUC-ROC (image level) | AUC-PR (subject level) | AUC-ROC (subject level) |
|---|---|---|---|---|
| PIV ≥ 0 | | | | |
| Color RFI | 0.999 | 0.999 | 1.000 | 1.000 |
| Grayscale RVM | 0.938 | 0.959 | 0.995 | 0.995 |
| Binarized RVM | 0.960 | 0.974 | 0.999 | 0.999 |
| Skeletonized RVM | 0.882 | 0.944 | 0.980 | 0.990 |
| PIV ≥ 50 | | | | |
| Grayscale RVM | 0.844 | 0.891 | 0.978 | 0.984 |
| Binarized RVM | 0.846 | 0.897 | 0.986 | 0.988 |
| Skeletonized RVM | 0.809 | 0.872 | 0.972 | 0.970 |
| PIV ≥ 100 | | | | |
| Grayscale RVM | 0.809 | 0.869 | 0.976 | 0.982 |
| Binarized RVM | 0.810 | 0.869 | 0.950 | 0.968 |
| Skeletonized RVM | 0.762 | 0.837 | 0.943 | 0.959 |
| PIV ≥ 150 | | | | |
| Grayscale RVM | 0.773 | 0.840 | 0.933 | 0.965 |
| Binarized RVM | 0.781 | 0.848 | 0.969 | 0.972 |
| Skeletonized RVM | 0.756 | 0.829 | 0.945 | 0.961 |
| PIV ≥ 200 | | | | |
| Grayscale RVM | 0.710 | 0.790 | 0.929 | 0.936 |
| Binarized RVM | 0.679 | 0.775 | 0.839 | 0.891 |
| Skeletonized RVM | 0.705 | 0.795 | 0.902 | 0.943 |
| PIV ≥ 210 | | | | |

| | | | | | |
|---|---|---|---|---|---|
| | Grayscale RVM | 0.624 | 0.725 | 0.865 | 0.894 |
| | Binarized RVM | 0.678 | 0.778 | 0.835 | 0.859 |
| | Skeletonized RVM | 0.665 | 0.767 | 0.832 | 0.882 |
| PIV ≥ 220 | | | | | |
| | Grayscale RVM | 0.562 | 0.706 | 0.667 | 0.778 |
| | Binarized RVM | 0.638 | 0.752 | 0.840 | 0.846 |
| | Skeletonized RVM | 0.611 | 0.724 | 0.863 | 0.866 |
| PIV ≥ 230 | | | | | |
| | Grayscale RVM | 0.540 | 0.679 | 0.669 | 0.748 |
| | Binarized RVM | 0.518 | 0.654 | 0.729 | 0.764 |
| | Skeletonized RVM | 0.488 | 0.639 | 0.590 | 0.703 |
| PIV ≥ 240 | | | | | |
| | Grayscale RVM | 0.450 | 0.597 | 0.669 | 0.716 |
| | Binarized RVM | 0.463 | 0.608 | 0.504 | 0.655 |
| | Skeletonized RVM | 0.451 | 0.606 | 0.549 | 0.677 |
| PIV ≥ 250 | | | | | |
| | Grayscale RVM | 0.406 | 0.528 | 0.515 | 0.654 |
| | Binarized RVM | 0.394 | 0.515 | 0.382 | 0.505 |
| | Skeletonized RVM | 0.397 | 0.522 | 0.394 | 0.526 |
| PIV ≥ 256 | | | | | |
| | Grayscale RVM | 0.390 | 0.504 | 0.397 | 0.500 |
| | Binarized RVM | 0.390 | 0.504 | 0.397 | 0.500 |
| | Skeletonized RVM | 0.390 | 0.504 | 0.397 | 0.500 |
| PIV ≤ 10 | | | | | |
| | Grayscale RVM | 0.851 | 0.905 | 0.945 | 0.965 |

| | | | | |
|---|---|---|---|---|
| Binarized RVM | 0.949 | 0.970 | 0.997 | 0.997 |
| Skeletonized RVM | 0.913 | 0.949 | 0.994 | 0.996 |
| 75 ≥ PIV ≥ 150 | | | | |
| Grayscale RVM | 0.807 | 0.876 | 0.972 | 0.978 |
| Binarized RVM | 0.807 | 0.872 | 0.973 | 0.983 |
| Skeletonized RVM | 0.773 | 0.848 | 0.972 | 0.980 |

Abbreviations: PIV – pixel intensity values, RFI – retinal fundus image, RVM – retinal vessel maps, AUC-PR – area under the precision recall curve, AUC-ROC – area under the receiver operating characteristics curve

**Note**: The null AUC-PR is ~0.387 at the image level and ~0.386 at the subject level.

**Supplemental Table 2: Performance of models trained by PS.**

| IMAGE TYPE | AUC-PR (image level) | AUC-ROC (image level) | AUC-PR (patient level) | AUC-ROC (patient level) |
|---|---|---|---|---|
| Color RFI | 0.98 | 0.98 | 0.99 | 0.99 |
| Grayscale RVM | 0.97 | 0.97 | 0.96 | 0.96 |

## SUPPLEMENTAL METHODS

### *Experiment Reproduction by PS*

*Dataset*

To further validate our hypothesis, a subset of the experiments were independently repeated by a different researcher (PS) by partitioning the original dataset into different training, validation and test sets as well as by adopting a different training schedule. To this end, the original dataset, consisting of 262 babies (99Black/163 White) and 4546 RFIs (1709 Black/2837 White), was partitioned (at the patient level) into train, validation and test sets with a 60:20:20 split, respectively. After splitting, 156 babies (54 Black/102 White) & 2764 RFIs (994 Black/1770 White) were in the training dataset, 53 babies (21 Black/32 White) & 973 RFI's (392 Black/581 White) were in the validation set, and 53 babies (24 Black/29 White) and 809 RFIs (323 Black/486 White) were in the test set.

*Model*

The binary race classification model used an Imagenet pretrained Resnet18 deep learning architecture for training on the above dataset.[37] As a loss function, binary cross entropy was used. The original last fully connected layer in the pre-trained model was converted to generate a single neuron output (to predict whether a sample belongs to class "White" or not). Owing to imbalance in Black & White samples in the train set (35.96% Black, 64.04% White), a weighted sampler that draws more black samples than white samples was utilized.

For better generalization, each RFI/RVM was transformed with random rotation (with probability p=0.1), horizontal & vertical flip (p=0.5) as well as random zoom-in (p=0.5). All RFI/RVM's were resized to a standard 640 X 480 size before passing through the network. For improved visibility of vascular structures, RFI images were additionally preprocessed with CLAHE after converting from RGB images to LAB color space (and finally switching back to RGB before inputting them to the model). The DL model was trained for 30 epochs, using the Adam optimizer with a learning rate of 0.0001 and batch size of 64.[38] To prevent overfitting on the train set, early stopping was implemented, with the best model selected via the AUC-ROC score on the validation set. While the Resnet18 model, loss function, and data loader packages were imported from Pytorch DL framework, we used the various transformation functions as well as the performance metric using Monai.[39,40]